\documentclass{article}
\usepackage{arxiv} % arxiv style
\usepackage[hyphens]{url} % for urls with proper breaking
\usepackage[utf8]{inputenc} % allow utf-8 input
\usepackage{amsmath} % for maths
\usepackage{amssymb} % for extra maths symbols
\usepackage[a-1b]{pdfx} % for accesibility of pdf (loads hyperref)
\hypersetup{hidelinks} % option for hyperref
\usepackage{xcolor} % for colors
\usepackage[normalem]{ulem} % for text highlighting compatible with accesibility
\usepackage{microtype} % improves typography
\usepackage{algorithm, algorithmicx, algpseudocode} % algorithm
\usepackage{graphicx} % to insert pictures
\usepackage{caption, subcaption} % for subfigures

\newcommand\hl{\bgroup\markoverwith{\textcolor{yellow}{\rule[-.5ex]{2pt}{2.5ex}}}\ULon}

\title{Ranger21: a synergistic deep learning optimizer}

\author
{
  Less~Wright \\
  AudereNow.org \\
  1191 2nd Ave Ste 450\\
  Seattle, WA 98101 USA \\
  \texttt{less@auderenow.org} \\
\And
  Nestor~Demeure\\
  National Energy Research Scientific Computing Center, \\
  Lawrence Berkeley National Lab,\\
  1 Cyclotron Road, Berkeley, California, 94720 \\
  \texttt{ndemeure@lbl.gov} \\
}

\begin{document}
\maketitle

% abstract
\begin{abstract}
%Sentence 1: State the problem
As optimizers are critical to the performances of neural networks, every year a large number of papers innovating on the subject are published.
%Sentence 2: State the consequences of the problem
However, while most of these publications provide incremental improvements to existing algorithms, they tend to be presented as new optimizers rather than composable algorithms.
Thus, many worthwhile improvements are rarely seen out of their initial publication.
%Sentence 3: State your solution
Taking advantage of this untapped potential, we introduce Ranger21, a new optimizer which combines AdamW with eight components, carefully selected after reviewing and testing ideas from the literature.
%Sentence 4: State the consequences of the solution
We found that the resulting optimizer provides significantly improved validation accuracy and training speed, smoother training curves, and is even able to train a ResNet50 on ImageNet2012 without Batch Normalization layers.
A problem on which AdamW stays systematically stuck in a bad initial state.
\end{abstract}

% keywords
\keywords{Deep-learning \and Optimizer}

% body of the text
\section{Introduction}

% Context, why it matters
Once a task has been selected, an architecture designed and a dataset assembled, one needs to train a neural network in order to make it performant as most neural network return random outputs in the absence of training.
Thus, it has long been understood that optimizers are a primordial component of deep-learning and that a good optimizer can drastically improve the performance of a given architecture.

Currently Adam \cite{ref_Adam}, which is short for Adaptive Moment Estimation, and in particular its variant AdamW \cite{ref_AdamW} are among the most commonly used optimizers in deep learning \cite{ref_adampopular}.
As such, many papers have proposed and tested new optimization ideas that modify or enhance the core adaptive moment estimation optimization algorithm in various aspects including forms of gradient normalization, improvement to the computation of the momentum, to the computation of the weight decay and of the step-size.
% Our solution / contributions
We have found that these ideas, while typically presented as new optimizers, are often composable.
When tested and carefully combined, this can produce a synergistic result, where the combined result is stronger than the individual components alone. 
Ranger21 is our attempt at building such a synergistic optimizer by selecting, testing and integrating eight distinct optimizer enhancements on top of AdamW.

Our work derives from Ranger \cite{ref_ranger}, which was built in 2019 by combining two such algorithms (Rectified Adam \cite{ref_RAdam} and Lookahead \cite{ref_lookahead}) into a single optimizer.
As the name implies, Ranger21 synthesize our work to test and integrate new algorithms through 2021.
% summary of results
When testing our optimizer on ImageNet2012 \cite{ref_imagenet}, as seen in section \ref{sec:experiments}, we found out that it provided consistent improvements over AdamW.
Training was smoother, achieved certain validation levels up to 3x faster, and most importantly, achieved a higher final validation accuracy.
Furthermore, it was even able to train an architecture that AdamW completely fails at training (a ResNet50 \cite{ref_resnet} without its batch-normalization layers).

% organization of the paper
This publication is split into five sections.
Section one is this introduction, section two details the components that makes Ranger21 while section three gives the details of the full algorithm.
Section four is dedicated to our experiments on the ImageNet2012 dataset and section five concludes with our results and potential improvements.

\section{Components of the optimizer}

This section introduces the various components that make our optimizer and describes them as they are implemented in Ranger21.

%\begin{enumerate}
%    \item Adam optimizer
%    \item Adaptive Gradient Clipping
%    \item Gradient Centralization
%    \item Positive-Negative Momentum
%    \item Norm Loss
%    \item Stable weight decay
%    \item Linear learning rate warm-up
%    \item Explore-Exploit learning rate schedule
%    \item Lookahead
%\end{enumerate}

\subsection{Adam (adaptive moment estimation)}

Adam \cite{ref_Adam} and in particular AdamW \cite{ref_AdamW}, which is detailed in algorithm \ref{alg:adam}, is the core of the Ranger21 optimizer.
Adam is short for Adaptive Moment Estimation and, as the name implies, it computes an adaptive step size for each parameter by maintaining two exponential moving averages which track the first and second moment of the stochastic gradients.  

By using the sign of the stochastic gradient and a computed estimate of relative (or adaptive) variance derived from the two moving averages, the magnitude and direction of the update are updated at each optimizer step.

\begin{algorithm}[h]
\caption{AdamW}
\label{alg:adam}
\begin{algorithmic}[1] % [1] to get line numbers
  \Require $f(\theta)$: objective function
  \Require $\theta_0$: initial parameter vector
  \Require $\eta$: learning rate
  \Require $\lambda$: weight decay (default: $1e^{-4}$)
  \Require $\beta_1, \beta_2 \in \left[ 0, 1\right[$: decay rates for Adam (default: $0.9, 0.999$)
  \Require $\epsilon$: epsilon for numerical stability (default: $1e^{-8}$)
  \Require $t_{max}$: number of iterations
  \State $m_0, v_0 \gets 0, 0$ \Comment{Initialization}
  \For{$t \gets 1$ to $t_{max}$}
    \State $g_t \gets \nabla f_{t}(\theta_{t-1})$ \Comment{Gradient}
    \State $m_t \gets \beta_1\,m_{t-1} + (1-\beta_1)\,g_t$ \Comment{1st mom. estimate} \label{alg:adam:1mom}
    \State $\widehat{m}_t \gets m_t / (1 - \beta_1^t)$ \Comment{Bias correction}
    \State $v_t \gets \beta_2\,v_{t-1} + (1-\beta_2)\,g_t^2$ \Comment{2nd mom. estimate}
    \State $\widehat{v}_t \gets v_t / (1 - \beta_2^t)$ \Comment{Bias correction} \label{alg:adam:2bias}
    \State $u_t \gets \widehat{m}_t / (\sqrt{\widehat{v}_t} + \epsilon)$ \Comment{Update vector}
    \State $d_t \gets \lambda \theta_{t-1}$ \Comment{Weight decay}
    \State $\theta_t \gets \theta_{t-1} - \eta u_t - \eta d_t$ \Comment{Apply parameter update}
  \EndFor \\
  \Return $\theta_t$
\end{algorithmic}
\end{algorithm}

Due to Adam(W) being one of the most frequently used optimizers, \cite{ref_adampopular}, many publications offer various incremental and innovative improvements on the algorithm, which is of particular interest to us as these individual enhancements are often composable.

\subsection{Adaptive Gradient Clipping}
% https://arxiv.org/abs/1905.11881v2  (theory of gradient clipping, smooths the loss landscape)
% https://arxiv.org/abs/2102.06171v1  (unit wise agc - what we use in ranger 21)

Sporadic 'high loss' mini-batches can destabilize stochastic gradient descent due to the back-propagation of excessively large gradients. This is a common issue with smaller batch sizes and higher learning rates.
To solve this problem one can use gradient-clipping, ensuring that the gradient stays below a given threshold (as seen in equation \ref{eq:gradient_clipping} where $\tau$ is the threshold and the norm is computed with a Frobenius norm).

\begin{equation}
\label{eq:gradient_clipping}
    g_t =
    \begin{cases}
    {\color{blue} \tau} \frac{ g_t }{\color{blue} \|g_t\| } & \text{if } \|g_t\| > \tau, \\
    g_t & \text{otherwise.}
    \end{cases}
\end{equation}

Theoretical studies have shown that gradient clipping should help the optimizer traverse non-smooth regions of the loss landscape and accelerate convergence \cite{ref_gradientclippingtheory}.
However, raw gradient clipping can affect the stability of the training, and finding a good threshold requires fine grained tuning based on the model depth, batch size and learning rate.

In Ranger21, we use Adaptive Gradient Clipping \cite{ref_adaptiveclipping}\footnote{We would like to thank Ali Kamer for pointing Adaptive Gradient Clipping to our attention.} (not to be confused with another method by the same name published previously in \cite{ref_adaptiveclipping_homonym}) in order to overcome these shortcomings.
 In Adaptive Gradient Clipping, the clipping threshold is dynamically updated to be proportional to the unit-wise ratio of gradient norms to parameter norms.  This is shown in equation \ref{eq:adaptive_gradient_clipping}. 
$\epsilon$ is a constant, set to $10^{-3}$ by default, added to avoid freezing zero-initialized parameters, $\tau$ is set to $10^{-2}$ by default and $r$ denotes the fact that we are working on individual rows of a layer rather than a full layer.

\begin{equation}
\label{eq:adaptive_gradient_clipping}
    g^r_t =
    \begin{cases}
    {\color{blue} \tau \frac{ max(\|\theta^r_t\|, \epsilon) }{ \|g^r_t\| } } g^r_t & \text{if } \frac{ \|g^r_t\| }{ max(\|\theta^r_t\|, \epsilon) } > \tau, \\
    g^r_t & \text{otherwise.}
    \end{cases}
\end{equation}

We found that Adaptive Gradient Clipping accelerates training while requiring only minimal tuning.

\subsection{Gradient Centralization}
% https://arxiv.org/abs/2004.01461

Inspired by normalization techniques such as Batch Normalization \cite{ref_batchnorm}, Gradient Centralization \cite{ref_gradientcentralization} suggests normalizing the gradient by subtracting its mean before using it inside an optimizer.

For each layer that has more than one dimension, the mean of the gradient is computed along all but the first axis (i.e. slice/matrix wise for convolution layers) and then subtracted from the gradient as shown in equation \ref{eq:gradient_centralization}.

\begin{equation}
\label{eq:gradient_centralization}
    g_{centralized_t} = \nabla f_{t}(\theta_{t-1}) {\color{blue} - mean\left( \nabla f_{t}(\theta_{t-1}) \right) }
\end{equation}

Gradient Centralization imposes a constraint on the loss function and acts as a regularizer which, according to the authors, should smooth the optimization landscape.
In practice, we observed improved generalization, a smoother training curve and faster convergence when using it on networks that contains either fully connected layers and/or convolutional layers.

\subsection{Positive-Negative Momentum}
% https://arxiv.org/abs/2103.17182

Momentum is used in modern deep learning optimizers to both smooth out the training noise and reduce the risk of the optimizer becoming stuck on saddle points and flat sections of the loss landscape.
The classical, Adam-style, formulation of momentum can be seen in lines \ref{alg:adam:1mom} to \ref{alg:adam:2bias} of algorithm \ref{alg:adam}.

Algorithm \ref{alg:positive_negative_momentum} illustrates Positive-Negative Momentum (and in particular AdaPNM), a variant introduced in \cite{ref_posnefmomentum}.
Notice that the update vector is normalized by $\sqrt{(1 + \beta_0 )^2 + \beta_0^2}$ such that changing the momentum hyper-parameter does not require updating the learning rate.

\begin{algorithm}[h]
\caption{Positive-Negative Momentum}
\label{alg:positive_negative_momentum}
\begin{algorithmic}[1] % [1] to get line numbers
    \Require $g_t$: gradient of the objective function
    \Require ${\color{blue} \beta_0}, \beta_1, \beta_2 \in \left[ 0, 1\right[$: decay rates for the momentums (default: $0.9, 0.9, 0.999$)
    \Require $\epsilon$: epsilon for numerical stability (default: $1e^{-8}$)
    \State $m_t \gets \beta_1^{\color{blue} 2} m_{\color{blue} t-2} + (1-\beta_1^{\color{blue} 2}) g_t$ \Comment{1st mom. estimate}
    \State $\widehat{m}_t \gets {\color{blue} ( (1 +  \beta_0) m_t  - \beta_0 m_{t-1} )} / (1 - \beta_1^t) $ \Comment{Positive-negative momentum and bias correction}
    \State $v_t \gets \beta_2 v_{t-1} + (1 - \beta_2) g_t^2$ \Comment{2nd mom. estimate}
    \State ${\color{blue} v_{\mathrm{max}} \gets \max(v_t, v_{\mathrm{max}}) }$ \Comment{2nd mom. maximum estimate}
    \State $\widehat{v}_t \gets {\color{blue} v_{\mathrm{max}}} / (1-\beta_2^t) $ \Comment{Bias correction}
    \State $u_t \gets \widehat{m}_t / ( {\color{blue} \sqrt{(1 + \beta_0 )^2 + \beta_0^2} }(\sqrt{\widehat{v}_t} + \epsilon) )$ \Comment{Update vector}
\end{algorithmic}
\end{algorithm}

The key idea of Positive-Negative Momentum is to keep two sets of first moment estimates, one for odd iterations and one for even iterations.
The moment applied during the optimization is an average of both sets, assigning a positive weight to the current momentum estimation and a negative step to the previous one.

According to \cite{ref_posnefmomentum}, this simulates the addition of a parameter-dependent, anisotropic, noise to the gradient which should help escaping saddle points and pushing the optimizer toward flatter minima which are theorized to lead to better generalization.

In our tests, we were able to verify experimentally that Positive-Negative Momentum does indeed lead to improved performances on a variety of datasets, and integrates in a complementary fashion with the additional algorithms used in Ranger21.

\subsection{Norm loss}
% https://arxiv.org/abs/2103.06583

In AdamW style optimizers, weight decay is computed as in equation \ref{eq:adamw_decay} (where $\eta$ is the learning rate, $\lambda$ the parameter scaling the weight decay and $\theta$ the parameters that we are optimizing) and subtracted from the parameters during the update step.

\begin{equation}
\label{eq:adamw_decay}
    d_{AdamW} = - \eta \lambda \theta
\end{equation}

Norm Loss \cite{ref_normloss} suggests using equation \ref{eq:normloss_decay}.
Given a weight matrix $c_o$ (which could be associated with a linear layer or a convolution layer for example), it takes into account $\|\theta_{c_o}\|$, the euclidian norm of the weight matrix, such that the weight matrix is pushed toward a unit norm unlike traditional weight decay, which consistently pushes the weights toward zero.

\begin{equation}
\label{eq:normloss_decay}
    d_{NormLoss_{c_o}} = - \eta \lambda {\color{blue} \left( 1 - \frac{1}{\|\theta_{c_o}\|} \right) } \theta
\end{equation}

In testing, we found that Norm Loss acts as a soft regularizer for the weight space and performs well across a large variety of hyper-parameters.

\subsection{Stable Weight Decay}
% https://arxiv.org/abs/2011.11152

AdamW-style weight decay uses the learning rate of the optimizer to weight the decay, as seen in equation \ref{eq:adamw_decay}.
However, the actual step size is not only a function of the learning rate but also of $\widehat{v}_t$, which represents the magnitude of the gradient.
Thus, the actual step size evolves during the iterations and a weight decay calibrated for the first iterations of training will be too large for later iterations when $\widehat{v}_t$ goes down towards zero.

To solve this problem, \cite{ref_stableweightdecay} introduces Stable Weight Decay which, as seen in equation \ref{eq:stable_decay} (where \textit{mean} is the average across all coefficients of a layer), is a reweighting of AdamW-style weight decay to take the effective step size into account.

\begin{equation}
\label{eq:stable_decay}
    d_{stable} = - \frac{\eta}{\color{blue} \sqrt{mean(\widehat{v}_t)}} \lambda \theta
\end{equation}

In our tests, we found that Stable Weight Decay gave us significant generalization improvements on vision tasks, in keeping with the authors affirmation that Stable Weight Decay allows adaptive optimizers to match and exceed SGD results on vision tasks.
Furthermore, we observed that it can be integrated seamlessly with Norm Loss (giving us equation \ref{eq:stableloss_decay}) and that the benefits of both methods are additive, as they approach weight regularization from different perspectives.

\begin{equation}
\label{eq:stableloss_decay}
    d_{stableloss} = - \frac{\eta}{\color{blue} \sqrt{mean(\widehat{v}_t)}} \lambda {\color{blue} \left( 1 - \frac{1}{\|\theta_{c_o}\|} \right) } \theta
\end{equation}

\subsection{Linear learning rate warm-up}
% https://arxiv.org/abs/1910.04209

The original Ranger optimizer was based on the Rectified Adam optimizer \cite{ref_RAdam} which tries to fix some of the instability problems that Adam encounters due to large updates in the first iterations.
However, \cite{ref_linearwarmup} subsequently introduced a much simpler alternative, relying only on a warm-up of the learning rate.
As can be seen in equation \ref{eq:lr_warmup}, we follow their suggestion and apply a linear warm-up of $2 \cdot (1 - \beta 2)^{-1}$ iterations to the learning rate (where $\beta_2$ is the second momentum parameter) which is about 2000 iterations for default hyperparameter values.
Since this schedule can produce a warm-up that is too long for shorter training runs, we additionally restrict it to the first  $t_{warmup}$ iterations (by default, $22\%$ of $t_{max}$, the total iterations).

\begin{equation}
\label{eq:lr_warmup}
    \eta_t = {\color{blue} 
    min\left( 1 , 
        max\left( \frac{1 - \beta_2}{2} \cdot t , \frac{t}{t_{warmup}} \right)
    \right) } \eta
\end{equation}

In our tests we found that this warm-up behave similarly to Rectified Adam, avoiding excessive step size during the first iterations, while being much simpler to implement.

\subsection{Explore-Exploit learning rate schedule}
% https://arxiv.org/abs/2003.03977
 
\cite{ref_kneeschedule} suggests an Explore-Exploit learning rate schedule: a phase of exploration, characterized by a constant high learning rate to find a flat minima, followed by a phase of exploitation where the learning rate is decreased linearly to zero.
The efficacy of this strategy has been confirmed experimentally in the Fastai leaderboard where the cosine annealing (which is extremely similar to the explore-exploit learning rate) dominates \cite[section 3]{ref_cosinefastai}.
As can be seen in equation \ref{eq:lr_warmdown}, given $t_{max}$ iterations, we decrease the learning-rate linearly during the last $t_{warmdown}$ iteration ($28\%$ of the iterations by default).

\begin{equation}
\label{eq:lr_warmdown}
    \eta_t = {\color{blue} min\left( 1 , \frac{t_{max} - t}{t_{warmdown}} \right) } \eta
\end{equation}

\begin{figure}[h]
\centering
\includegraphics[width=0.95\textwidth]{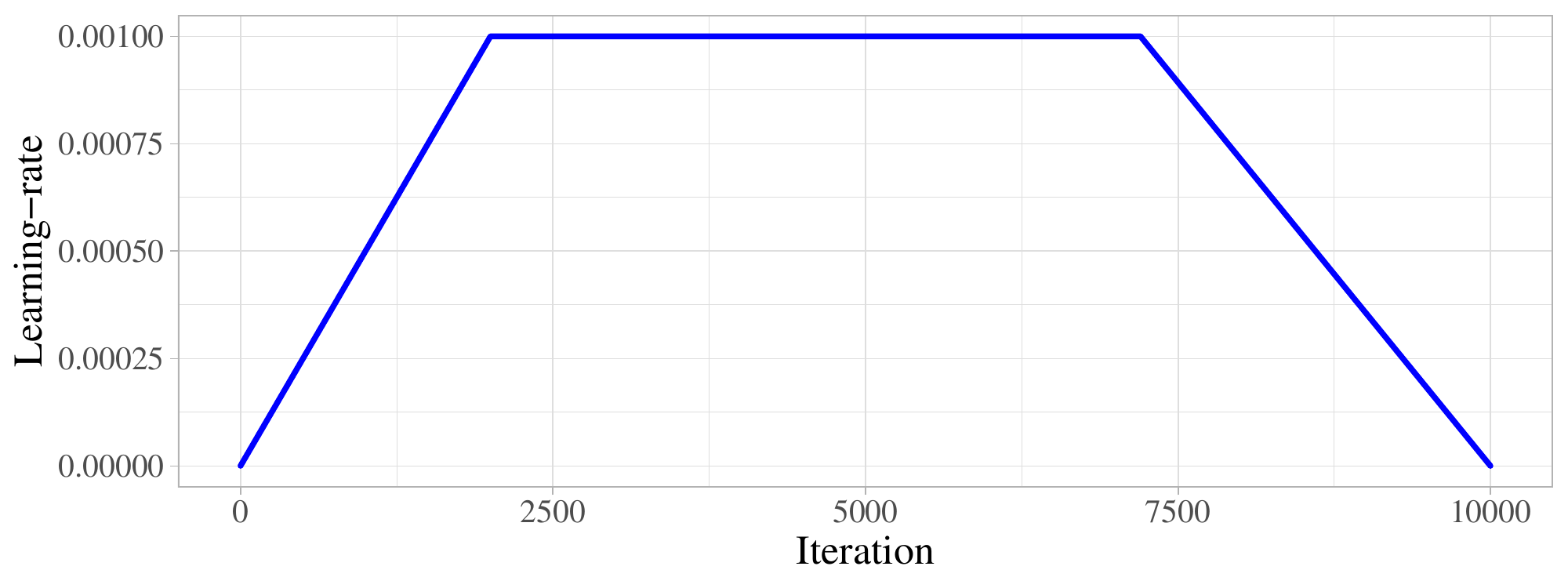}
\caption{Evolution of the learning rate accross $10000$ iterations with a base learning rate of $1e^{-3}$, a decay rate $\beta_2$ of $0.999$, a linear warm-up period of at most $22\%$ of the iterations and a linear warm-down period of $28\%$ of the iterations.}
\label{fig:learningrate_schedule}
\end{figure}

By combining the linear warm-up, a flat exploration phase, and a linear warm-down exploitation phase, we arrive at the learning-rate schedule given in equation \ref{eq:lr_fullschedule} and illustrated in figure \ref{fig:learningrate_schedule}.

\begin{equation}
\label{eq:lr_fullschedule}
    \eta_t = {\color{blue} min\left(1 , 
    max\left( \frac{1 - \beta_2}{2} \cdot t , \frac{t}{t_{warmup}} \right), 
    \frac{t_{max} - t}{t_{warmdown}} \right) } \eta
\end{equation}

\subsection{Lookahead}
% https://arxiv.org/abs/1907.08610

\cite{ref_lookahead} introduced Lookahead\footnote{Interestingly and in keeping with the spirit of this paper, the authors remark in the abstract of their publication that Lookahead is orthogonal to the majority of optimization techniques and can be used to improve most optimizers.}, a technique consisting of keeping an exponential moving average of the weights that is updated and substituted to the current weights every $k_{lookahead}$ steps (5 by default).
To implement Lookahead, one can apply algorithm \ref{alg:lookahead} at the end of the usual optimization step (where $\beta_{lookahead}$ is the momentum of the moving average, 0.5 by default).

\begin{algorithm}[h]
\caption{Lookahead}
\label{alg:lookahead}
\begin{algorithmic}[1] % [1] to get line numbers
    \Require $t$: current iteration number
    \Require $k_{lookahead}$: frequency of the update (default: $5$)
    \Require $\beta_{lookahead} \in \left[ 0, 1\right[$: decay rate (default: $0.5$)
    \State $l_0 \gets \theta_0$ \Comment{Initialization}
    \State ... \Comment{Usual parameter update}
    \If{t \% $k_{lookahead}$ == 0}
        \State $l_{t/k} \gets \beta_{lookahead} l_{t/k-1} + (1 - \beta_{lookahead}) \theta_t$ \Comment{Apply linear interpolation}
        \State $\theta_t \gets l_{t/k}$ \Comment{Apply parameter update}
    \EndIf{}
\end{algorithmic}
\end{algorithm}

Ranger, our first optimizer which has now evolved into Ranger21, was built on a mix of the Rectified Adam optimizer and Lookahead.
While designing Ranger21, we found that Lookahead's benefit subsists through the additions to the algorithm and that it gives us a net upward shift in the training validation curve on various datasets and a higher final validation accuracy.

\section{The Ranger21 optimizer}

Algorithm \ref{alg:ranger21} introduces Ranger21\footnote{It gets its name from the current year, 2021, and the fact that it evolved organically from the Ranger optimizer.} which is the combination of all the aforementioned components.

\begin{algorithm}[h]
\caption{Ranger21}
\label{alg:ranger21}
\begin{algorithmic}[1] % [1] to get line numbers
  \Require $f(\theta)$: objective function
  \Require $\theta_0$: initial parameter vector
  \Require $\eta$: learning rate
  \Require $\lambda$: weight decay (default: $1e^{-4}$)
  \Require $\beta_0, \beta_1, \beta_2, \beta_{lookahead} \in \left[ 0, 1\right[$: decay rates (default: $0.9, 0.9, 0.999, 0.5$)
  \Require $\epsilon, \epsilon_{clipping}$: epsilon for numerical stability (default: $1e^{-8}, 1e^{-3}$)
  \Require $\tau_{clipping}$: threshold for adaptive gradient clipping (default: $10^{-2}$ )
  \Require $k_{lookahead}$: frequency of the update (default: $5$)
  \Require $t_{max}$: number of iterations
  \Require $t_{warmup}$: number of learning rate warm-up iterations (default: $0.22 \times t_{max}$)
  \Require $t_{warmdown}$: number of learning rate warm-down iterations (default: $0.28 \times t_{max}$)
  \State $m_0, v_0, v_{\mathrm{max}} \gets 0, 0, 0$ \Comment{Initialization}
  \State $l_0 \gets \theta_0$ \Comment{Lookahead initialization}
  \For{$t \gets 1$ to $t_{max}$}
    \State $g_t \gets \nabla f_{t}(\theta_{t-1})$ \Comment{Gradient}
    \For{$r \in rows(g_t)$}
        \If{$\frac{ \|g^r_t\| }{ max(\|\theta^r_t\|, \epsilon_{clipping}) } > \tau_{clipping}$}
            \State $g^r_t \gets \tau_{clipping} \frac{ max(\|\theta^r_t\|, \epsilon_{clipping}) }{ \|g^r_t\| } g^r_t$ \Comment{Gradient clipping}
        \EndIf{}
    \EndFor 
    \State $g_t = g_t - mean\left( g_t \right)$ \Comment{Gradient centralization}
    \State $m_t \gets \beta_1^2 m_{t-2} + (1-\beta_1^2) g_t$ \Comment{1st mom. estimate}
    \State $\widehat{m}_t \gets ( (1 +  \beta_0) m_t  - \beta_0 m_{t-1} ) / (1 - \beta_1^t) $ \Comment{Bias correction}
    \State $v_t \gets \beta_2 v_{t-1} + (1 - \beta_2) g_t^2$ \Comment{2nd mom. estimate}
    \State $v_{\mathrm{max}} \gets \max(v_t, v_{\mathrm{max}})$ \Comment{2nd mom. maximum estimate}
    \State $\widehat{v}_t \gets v_{\mathrm{max}} / (1-\beta_2^t) $ \Comment{Bias correction}
    \State $u_t \gets \widehat{m}_t / ( \sqrt{(1 + \beta_0 )^2 + \beta_0^2 }(\sqrt{\widehat{v}_t} + \epsilon) )$ \Comment{Update vector}
    \State $\eta_t = min\left(1 , max\left( \frac{1 - \beta_2}{2} \cdot t , \frac{t}{t_{warmup}} \right), \frac{t_{max} - t}{t_{warmdown}} \right) \eta$ \Comment{Learning rate scheduling} 
    \State $d_t = \frac{\eta_t}{\sqrt{mean(\widehat{v}_t)}} \lambda \left( 1 - \frac{1}{\|\theta_{t-1}\|} \right) \theta_{t-1}$ \Comment{Weight decay}
    \State $\theta_t \gets \theta_{t-1} - \eta_t u_t - \eta_t d_t$ \Comment{Parameter update}
    \If{t \% $k_{lookahead}$ == 0}
        \State $l_{t/k} \gets \beta_{lookahead} l_{t/k-1} + (1 - \beta_{lookahead}) \theta_t$
        \State $\theta_t \gets l_{t/k}$ \Comment{Lookahead}
    \EndIf{}
  \EndFor \\
  \Return $\theta_t$
\end{algorithmic}
\end{algorithm}

While the algorithm can appear daunting, it can be reduced to the eight previously mentioned ideas applied to the Adam optimizer.
Ranger21 combines these based on the findings that they are orthogonal, compatible with the Adam optimizer and, in our tests, synergistic.

It is interesting to note that while the algorithm adds a number of hyper-parameters, in our experience and across a wide range of tasks, the default values for those additional parameters perform well without tuning.  
This is, in part, due to the fact that some algorithms we included (such as Stable Weight Decay) strive to reduce the sensitivity of the algorithm to it's hyper-parameters.

\section{Experiments}
\label{sec:experiments}

Rather than using toy datasets, whose results often do not transfer to real world usage, we decided to focus on the ImageNet2012 dataset \cite{ref_imagenet} to demonstrate the accuracy of our optimizer.
To do so, we trained networks using both Ranger21 and AdamW for 60 epochs\footnote{Training on ImageNet2012 is often done in 300 epochs but it is computationally expensive and brings very little added value to this comparison.} with a default learning rate of $3e^{-3}$.

We used pictures of size 288x288 that were resized and either randomly cropped to 256x256 for the training set or center cropped to 256x256 for the validation set.
To further augment the training set, we also introduce random flip along the horizontal axis and random modifications in brightness, contrast and saturation (all of which are common data augmentation techniques).

We used Label Smoothing \cite{ref_labelsmoothing} (with a smoothing parameter of 0.1) as our loss function. It is commonly introduced to enhance generalization by improving the model's expected calibration error.

\subsection{ResNet50}

\begin{figure}[h]
\centering
\begin{subfigure}[t]{0.4\textwidth} % 7 / 17
    \centering
    \includegraphics[width=\textwidth]{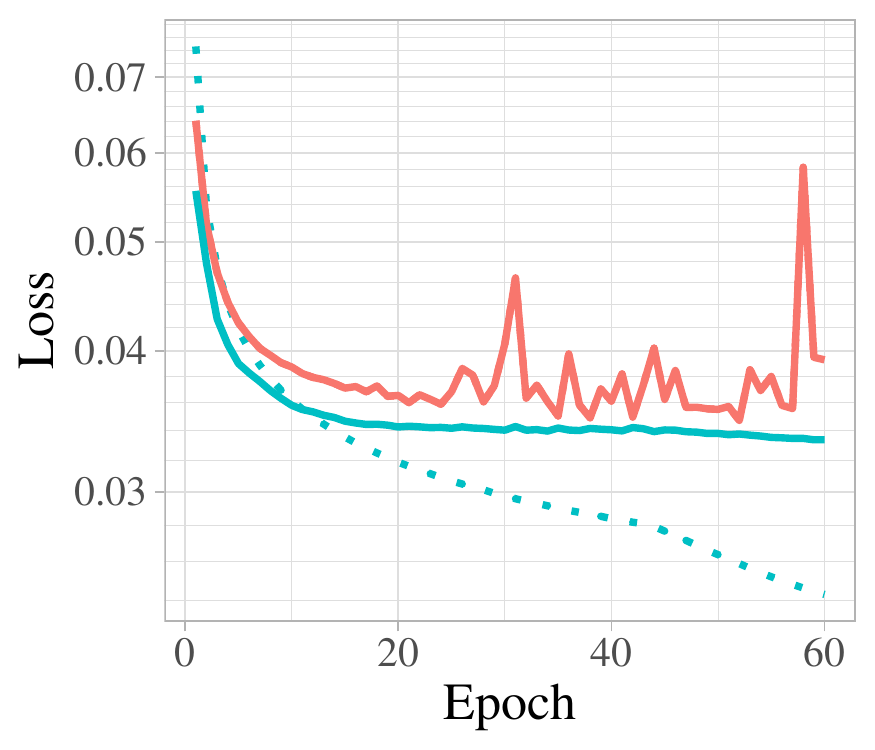}
    %\caption{Loss.}
    \label{fig:resnet50_loss}
\end{subfigure}
\hfill
\begin{subfigure}[t]{0.57\textwidth} % 10 / 17
    \centering
    \includegraphics[width=\textwidth]{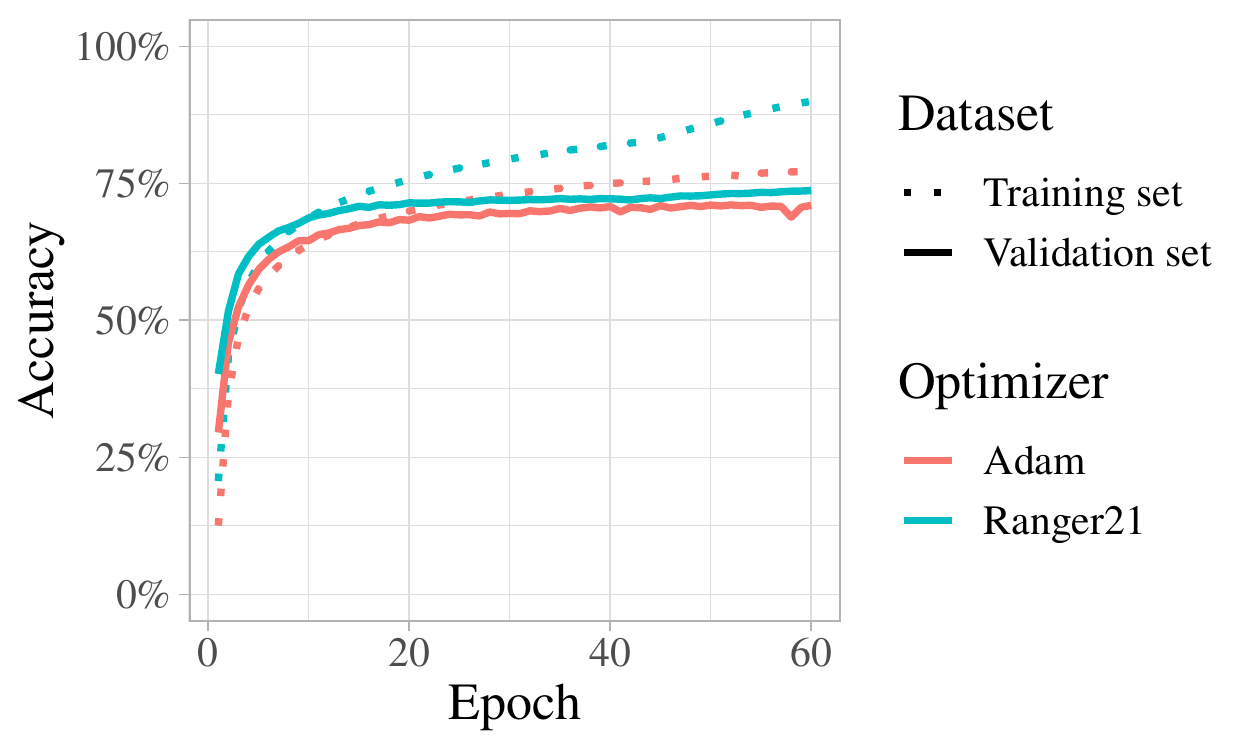}
    %\caption{Accuracy.}
    \label{fig:resnet50_acc}
\end{subfigure}
\caption{Evolution of the loss and accuracy, using either Adam or Ranger21, when training a ResNet50 convolutional neural network on the ImageNet dataset.
The y axis of the loss is displayed in logarithmic scale while the y axis of the accuracy is in percent.}
\label{fig:resnet50}
\end{figure}

First, we trained a very classical ResNet50 architecture \cite{ref_resnet}.
At its highest point (the final iteration), Ranger21 reaches a validation accuracy of $73.69\%$ which is $2.61\%$ higher than AdamW's highest accuracy ($71.08\%$, reached on epoch 52).
Furthermore, as can be seen in picture \ref{fig:resnet50} (and in particular the loss section), Ranger21 leads not only to a consistently better loss, but also to a much smoother training curve (which partially explains why the best result was reached on the final iteration).

\begin{figure}[h]
\centering
\includegraphics[width=0.95\textwidth]{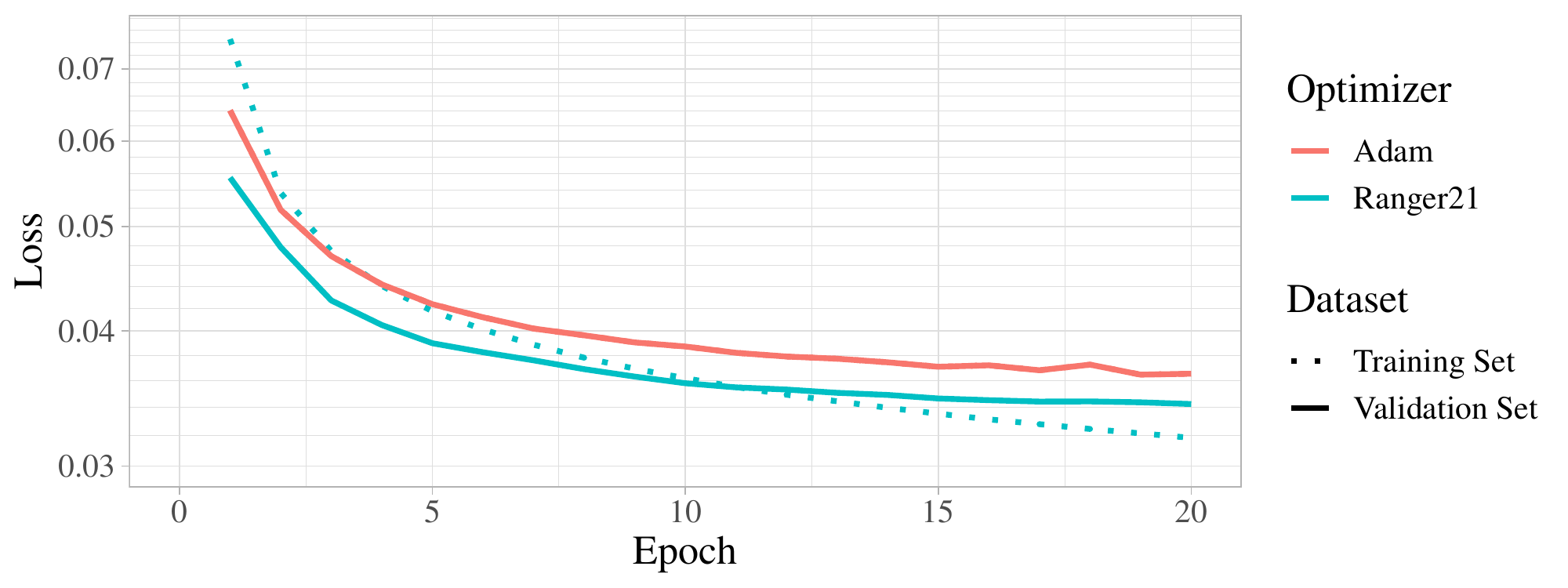}
\caption{Plot of the loss for the first 20 iterations, using either Adam or Ranger21, when training a ResNet50 convolutional neural network on the ImageNet dataset.
The y axis is displayed in logarithmic scale.}
\label{fig:zoomed_accuracy}
\end{figure}

The difference in convergence speed during training is well illustrated by the evolution of the loss during the first epochs of training, as seen in figure \ref{fig:zoomed_accuracy}. 
At 10 epochs Ranger21 reaches a result ($0.03579$) that will require AdamW more than 20 additional epochs to reach (it will only be equaled at epoch 35).

\subsection{Normalizer-Free ResNet50}

For our second experiment, we trained a Normalizer-Free ResNet50.
This architecture is identical to ResNet50 but omits all the batch normalization layers which previously helped smooth the loss landscape.

\begin{figure}[h]
\centering
\begin{subfigure}[t]{0.4\textwidth} % 7 / 17
    \centering
    \includegraphics[width=\textwidth]{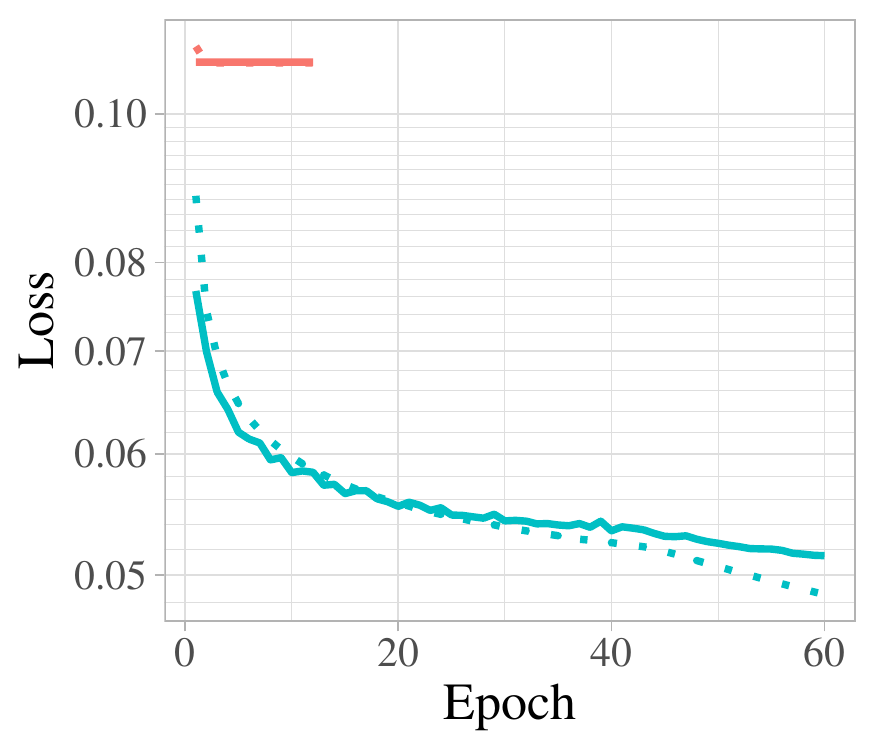}
    %\caption{Loss.}
    \label{fig:nfresnet50_loss}
\end{subfigure}
\hfill
\begin{subfigure}[t]{0.57\textwidth} % 10 / 17
    \centering
    \includegraphics[width=\textwidth]{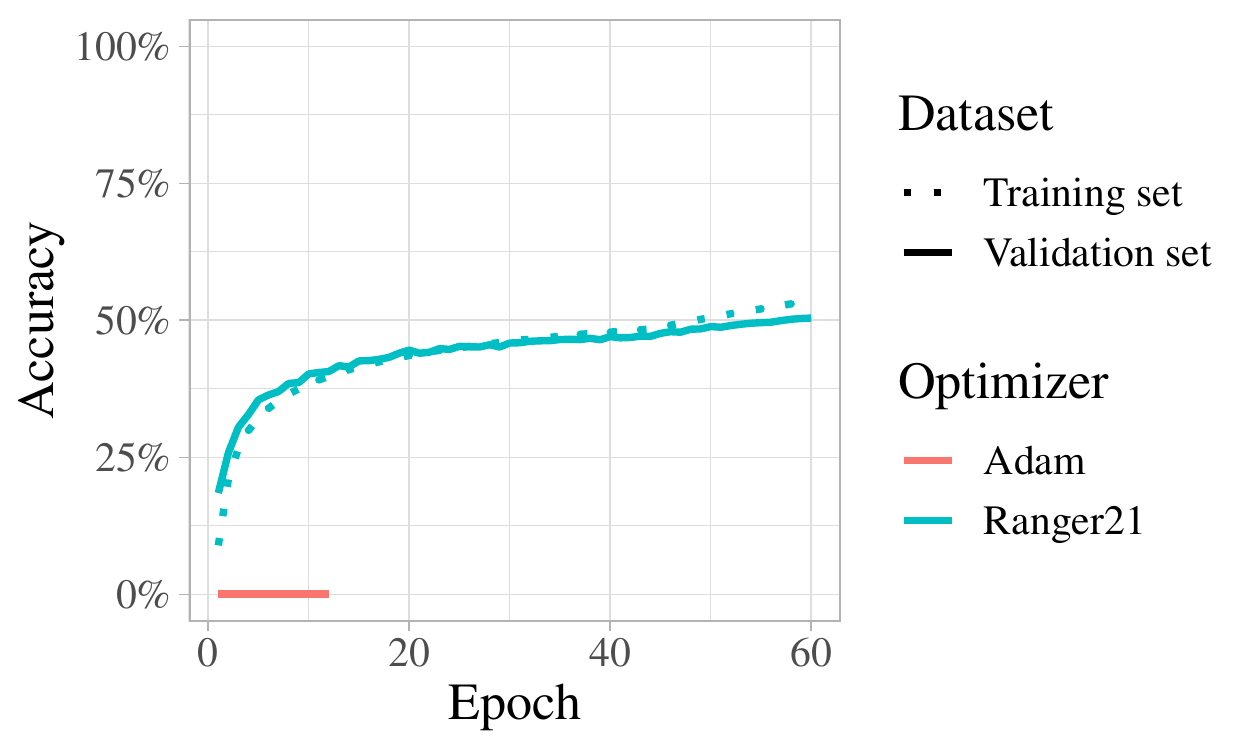}
    %\caption{Accuracy.}
    \label{fig:nfresnet50_acc}
\end{subfigure}
\caption{Evolution of the loss and accuracy, using either Adam or Ranger21, when training a Normalizer-Free ResNet50 convolutional neural network on the ImageNet dataset.
The y axis of the loss is displayed in logarithmic scale while the y axis of the accuracy is in percents.
Adam was stopped prematurely at 10 epochs as it was not converging.}
\label{fig:nfresnet50}
\end{figure}

As can be seen in picture \ref{fig:nfresnet50}, AdamW is unable to converge for this architecture and keeps a constant, high ($0.1$), loss\footnote{We tried this experiment five times with different random seeds to ensure that is was not a stroke of particularly bad luck. All attempts failed similarly to the one shown.}.
By contrast, Ranger21 was able to train the network with slow (relative to a standard ResNet50) but steady loss improvements, and was still improving when we reached the 60 epoch cutoff point.

This is particularly interesting because it illustrate a profound difference in capacities between Ranger21 and AdamW, rather than an incremental improvement, despite Ranger21 deriving from AdamW.

\section{Conclusion}

% overview
Many publications introduce incremental improvements to existing optimizers, but present them as new optimizers, rather than as modules that could be combined.
We believe that being aware of this modularity is important in order to take full advantage of the research that is being done into deep learning optimization.
We designed Ranger21 to highlight the benefits that can be gained from such a combination: testing and combining multiple independent advancements into a singular optimizer that is significantly better than its individual parts.

By combining improvements in many sub-areas (such as momentum, loss and weight decay), we find that Ranger21 is able to train models that other optimizers simply fail to train, like a Normalizer-Free Resnet50.
More importantly, for a given model, Ranger21 is usually able to both accelerate the learning and achieve a net higher validation accuracy without compromising generalization.

While we have tested Ranger21 on several smaller datasets with even stronger out-performance, we felt that focusing on ImageNet2012 would be a good way to showcase the efficacy of our optimizer, as it is the gold standard in the domain of image classification\footnote{We also found that smaller datasets, such as MNIST \cite{ref_mnist} and CIFAR-10 \cite{ref_cifar10}, do not differentiate well between optimizers as all end up performing roughly equivalently due to the simplicity of the task.
Furthermore, testing on small scale datasets leads to results that are usually not representative of the results obtained when scaling to larger, real-world problems.}.
Thus, we hope that showing results on ImageNet2012 provides the reader with representative picture of our work, as it may apply to real-world datasets.

However, every dataset and network architecture creates a unique loss landscape and our experiments are only a proxy for the behaviour of Ranger21 in the wild.
Ranger21 is publicly available with ready to run PyTorch code\footnote{Available on Github at the following url: \url{https://github.com/lessw2020/Ranger21}} and Flax code\footnote{Available on Github at the following url: \url{https://github.com/nestordemeure/flaxOptimizers}} and we encourage users to try it for themselves and see if it brings real world improvement in their training\footnote{We welcome feedback to further refine the components that make Ranger21 and keep it up to date with the state of the art.}.

% perspective
Going forward, we are interested in trying to develop and integrate a finer-grained learning-rate scheduler to hopefully both improve performance and reduce the dependency on the default learning rate.

We have also experimented with transformers \cite{ref_transformer} type of architecture, following their success in vision applications, and would like to test whether some optimizer components are more beneficial to transformer than to other architectures.

Finally, a more general perspective would be the development of a fully modular optimizer architecture (possibly along the lines of Optax \cite{ref_optax}) coupled with a meta-optimizer (such as a combinatorial bandit algorithm \cite{ref_combinatorial_bandits}) that would pick and chose the components to build an optimizer tailor-made for a given task.

\section*{Acknowledgment}

This work was supported by the Director, Office of Science, Office of Advanced Scientific Computing Research, of the U.S. Department of Energy under Contract No. DE-AC02-05CH11231

% bibliography
\bibliographystyle{unsrt}  
\bibliography{bibliography}

\end{document}